%% file: main.tex
\documentclass{article}

\usepackage{microtype}
\usepackage{graphicx}
\usepackage{subcaption}
\usepackage{booktabs}
\usepackage{hyperref}

\usepackage[preprint]{icml2026}
\makeatletter
\renewcommand{\ICML@preprint}{\textit{Preprint. May 25, 2026. Accepted to the ICML 2026 Workshop on Hypothesis Testing.}}
\makeatother

\usepackage{amsmath}
\usepackage{amssymb}
\usepackage{mathtools}
\usepackage{amsthm}

\usepackage[capitalize,noabbrev]{cleveref}
\usepackage{enumitem}
\usepackage{array}
\usepackage{siunitx}
\usepackage{xcolor}

\theoremstyle{plain}

\theoremstyle{definition}

\theoremstyle{remark}

\input{macros}

\graphicspath{{figures/}}

\icmltitlerunning{Batch-Conditioned Refusal Testing}

\begin{document}

\twocolumn[
\icmltitle{A Paired Testing Protocol for\\
           Batch-Conditioned Refusal Robustness\\
           in LLM Serving}

\icmlsetsymbol{equal}{*}

\begin{icmlauthorlist}
\icmlauthor{Sahil Kadadekar}{ind}
\end{icmlauthorlist}

\icmlaffiliation{ind}{Independent Researcher}

\icmlcorrespondingauthor{Sahil Kadadekar}{sahilkadadekar@nyu.edu}

\icmlkeywords{hypothesis testing, LLM evaluation, refusal robustness, distribution shift, batch inference}
\hypersetup{
  pdftitle={A Paired Testing Protocol for Batch-Conditioned Refusal Robustness in LLM Serving},
  pdfauthor={Sahil Kadadekar},
  pdfsubject={Accepted to the ICML 2026 Workshop on Hypothesis Testing},
  pdfkeywords={hypothesis testing, LLM evaluation, refusal robustness, distribution shift, batch inference}
}

\vskip 0.3in
]

\printAffiliationsAndNotice{}

\begin{abstract}
Safety evaluations of language models often treat serving configuration as fixed background infrastructure, but batch condition is an untested treatment variable whenever the same prompt may be evaluated alone, in a synchronized batch, or inside a continuous-batching scheduler. We synthesize four artifact-backed studies into a paired testing protocol: Study A combines local discovery, scorer-corrected adjudication, and true-batching confirmation; Study B tests cross-model generalization; Study C tests continuous-batch composition; and Study D runs a batch-invariant-kernel ablation. The local test finds safety-label changes more often than capability-label changes (\baseSafetyFlip{} vs.\ \baseCapFlip{}), but adjudication of \trOneThirtyEightAdjudicatedRows{} candidate rows leaves only \trOneThirtyEightGenuineFlips{} genuine behavioral flips, implying a corrected full-set rate of \trOneThirtyEightCorrectedRate{}. The \mergedModels{}-model extension finds no detectable universal safety-over-capability skew: flips are near parity (\trOneFortyOneCombinedRatio{}), alignment type has no detectable association ($p=\trOneFortyOneAlignP{}$, $\eta^2=\trOneFortyOneAlignEta{}$), and output instability is the strongest tested fragility screen ($r=\trOneFortyOneInstabilityR{}$, bootstrap 95\% CI \trOneFortyOneInstabilityRCI{}). In the targeted kernel ablation, standard vLLM reproduces \kernelAblationStandardFlips{}/\kernelAblationRows{} label flips on current score-flip candidates, while enabling \texttt{VLLM\_BATCH\_INVARIANT=1} reduces the same test to \kernelAblationInvariantFlips{}/\kernelAblationRows{} flips; the composition test separately finds no aggregate effect at \trOneFortyThreeMDE{}pp sensitivity. The testing recommendation is exact-stack validation: evaluate refusal at the served batch setting, pair safety prompts with capability controls, and report low-rate directional flips separately from aggregate null effects.
\end{abstract}

\input{sections/01_introduction}
\input{sections/02_related_work}
\input{sections/03_experimental_setup}
\input{sections/04_results}
\input{sections/05_discussion_limitations}
\input{sections/06_conclusion}

\bibliographystyle{icml2026}
\bibliography{refs}

\appendix
\onecolumn
\input{sections/appendix_reproducibility}
\input{sections/appendix_audit_strengthening}
\input{sections/appendix_submission_materials}

\end{document}

%% file: macros.tex
\definecolor{MidnightBlue}{RGB}{0,51,102}

\newcommand{\trOneThirtyEightBase}{31{,}410}
\newcommand{\trOneThirtyEightRep}{7{,}257}
\newcommand{\trOneFortyOneTotal}{127{,}224}
\newcommand{\trOneFortyThreeTotal}{14{,}250}

\newcommand{\mergedModels}{15}

\newcommand{\baseSafetyFlip}{0.51\%}
\newcommand{\baseCapFlip}{0.14\%}
\newcommand{\repSafetyFlip}{1.68\%}
\newcommand{\repCapFlip}{0.42\%}
\newcommand{\trOneThirtyEightAuditUnsafe}{26}
\newcommand{\trOneThirtyEightAuditSafe}{18}
\newcommand{\trOneThirtyEightAdjudicatedRows}{63}
\newcommand{\trOneThirtyEightGenuineFlips}{17}
\newcommand{\trOneThirtyEightGenuinePct}{27}
\newcommand{\trOneThirtyEightArtifactPct}{73}
\newcommand{\trOneThirtyEightCorrectedRate}{0.16\%}
\newcommand{\trOneThirtyEightTrueBatch}{0.80\%}
\newcommand{\trOneThirtyEightTrueAgree}{99.4\%}

\newcommand{\kernelAblationRows}{55}
\newcommand{\kernelAblationRecords}{110}
\newcommand{\kernelAblationStandardFlips}{22}
\newcommand{\kernelAblationInvariantFlips}{0}
\newcommand{\kernelAblationStandardTextChanges}{25}
\newcommand{\kernelAblationInvariantTextChanges}{0}
\newcommand{\kernelAblationMatchedDirection}{15}

\newcommand{\kernelAblationSyncRows}{46}
\newcommand{\kernelAblationPromptListRows}{9}
\newcommand{\kernelAblationLlamaOneRows}{10}
\newcommand{\kernelAblationLlamaThreeRows}{19}
\newcommand{\kernelAblationQwenRows}{26}
\newcommand{\kernelAblationVllm}{0.19.1}

\newcommand{\trOneFortyOneCombinedRatio}{0.94\ensuremath{\times}}
\newcommand{\trOneFortyOneFragilityMin}{0.00\%}
\newcommand{\trOneFortyOneFragilityMax}{2.39\%}
\newcommand{\trOneFortyOneAlignP}{0.942}
\newcommand{\trOneFortyOneAlignEta}{0.033}
\newcommand{\trOneFortyOneAlignEquiv}{0.28pp}
\newcommand{\trOneFortyOneInstabilityR}{0.909}
\newcommand{\trOneFortyOneInstabilityRCI}{[0.65, 0.97]}
\newcommand{\trOneFortyOneInstabilityLOOCV}{[0.79, 0.93]}
\newcommand{\trOneFortyOneSafeDir}{159}
\newcommand{\trOneFortyOneUnsafeDir}{81}
\newcommand{\trOneFortyOnePhaseTwo}{0.80\%}
\newcommand{\trOneFortyOnePhaseAgree}{99.15\%}

\newcommand{\trOneFortyThreeUnsafeLow}{89}
\newcommand{\trOneFortyThreeUnsafeHigh}{92}
\newcommand{\trOneFortyThreePooledRatio}{90.3\%}
\newcommand{\trOneFortyThreePooledN}{28/31}
\newcommand{\trOneFortyThreePooledCI}{[75.1\%, 96.7\%]}
\newcommand{\trOneFortyThreeMixedP}{0.006}
\newcommand{\trOneFortyThreeCobatch}{22.1\%}
\newcommand{\trOneFortyThreeMDE}{4.7}

%% file: sections/01_introduction.tex
\section{Introduction}

Language-model evaluations usually hold serving configuration fixed.
That convention is convenient, but it hides a testing assumption: a refusal decision measured for a single request is treated as evidence about the same prompt under the served batch condition.
Batch size and scheduling are therefore not merely systems parameters when the outcome of interest is a boundary event such as refusal versus compliance.
They are treatment variables that should either be tested directly or explicitly ruled out.

This paper studies that testing problem for batch-conditioned refusal robustness.
The question is not whether batching is generally unsafe.
The question is whether a paired test can detect refusal-boundary movement when the same prompt is evaluated under different batch conditions, while controlling for ordinary output variation and benign capability changes.
The evidence is deliberately staged because the easy first conclusion is too strong.
An initial local test found more score-changing rows on the safety side than on the capability side, but adjudication sharply reduced the implied operational rate.
A \mergedModels{}-model extension found no universal safety-over-capability skew.
A continuous-batch composition study found no aggregate multi-tenant effect at the available sensitivity.
A targeted batch-invariant-kernel ablation then sharpened the mechanism story on the current candidate surface: standard vLLM reproduced \kernelAblationStandardFlips{}/\kernelAblationRows{} label flips, while the batch-invariant kernel produced \kernelAblationInvariantFlips{}/\kernelAblationRows{}.

We synthesize these studies into a compact testing protocol.
Study~A is a local paired perturbation test with safety prompts, capability controls, scorer correction, and a true-batching mechanism check.
Study~B scales the same question across \mergedModels{} models and asks whether model category or output instability predicts fragility.
Study~C tests whether continuous-batch co-residence creates a separate composition channel.
Study~D re-runs the current Study~A score-flip candidates under standard vLLM and vLLM's batch-invariant kernel on the same H100 serving stack.
Together, the studies separate discovery, measurement correction, generalization, mechanism confirmation, and deployment-composition testing.

\paragraph{Thesis.}
Batch-conditioned refusal flips exist, but they are low-rate and model-specific rather than a universal safety-over-capability law.
The correct output is therefore not a prohibition on batching.
It is an exact-stack test: evaluate the served batch setting, include capability controls, report scorer-corrected rates, and treat rare directional flips separately from aggregate null effects.

\paragraph{Scope posture.}
This is a workshop-scale synthesis across four artifact-backed studies rather than a new monolithic benchmark.
Its contribution is the testing contract that survives the staged evidence: what to compare, what to correct, what to generalize, and how not to over-claim a sparse-flip result.

\paragraph{Contributions.}
We make four testing contributions:
\begin{enumerate}[leftmargin=*]
  \item a paired protocol for batch-conditioned refusal testing that compares safety and capability labels under matched serving conditions and adds scorer-corrected adjudication;
  \item a cross-model generalization test showing that output instability predicts fragility more strongly than alignment type, while the aggregate safety-over-capability skew does not generalize;
  \item a continuous-batch composition test showing no aggregate effect at \trOneFortyThreeMDE{}pp sensitivity, despite rare directional flips leaning unsafe;
  \item a batch-invariant-kernel ablation showing that the current score-flip candidates disappear under the tested invariant vLLM execution path.
\end{enumerate}

%% file: sections/02_related_work.tex
\section{Related Work}

\subsection{Deterministic inference and batch-dependent variation}

The closest prior literature establishes the mechanism class but not the safety-facing claim. Work on deterministic inference shows that output variation can arise even when prompts, weights, and nominal decoding settings are fixed, because execution still depends on batching, scheduling, floating-point arithmetic, and backend kernels \citep{IEEE7542019,Atil2025DeterministicSettings,Gond2026LLM42,He2025DefeatingNondeterminism,LMSYS2025Deterministic}. The main concern in that literature is reproducibility: whether outputs can be made bitwise stable, what performance cost determinism imposes, and which execution paths trigger divergence. That work is directly relevant to our mechanism discussion, but it does not ask whether the resulting behavioral changes are concentrated in safety-sensitive rows.

Serving-systems work provides the architectural backdrop for why batching matters at all. Large-scale inference stacks now rely on tensor-parallel serving, iteration-level scheduling, continuous batching, key-value cache management, chunked prefill, and structured-generation runtimes to improve throughput and latency \citep{Aminabadi2022DeepSpeedInference,Yu2022Orca,Kwon2023PagedAttention,Agrawal2024Sarathi,Zheng2024SGLang}. These papers explain why batch composition, decode overlap, and memory behavior are first-order systems variables. We borrow that systems framing, but our target variable is different: we treat batch condition as a potential safety variable rather than only a performance variable.

\subsection{Behavioral sensitivity under deployment perturbations}

A second adjacent literature asks whether deployment choices preserve behavior that matters to downstream users. Broad evaluation frameworks already argue for measuring models across multiple risk dimensions rather than treating accuracy as sufficient \citep{Liang2023HELM,Wang2023DecodingTrust}. That principle is important for this paper because systems work tends to report throughput, latency, and memory, while alignment work tends to report refusal quality, jailbreak robustness, or harmful-completion rates. Very little work measures both on the same perturbation axis. That gap matters because batch effects are easy to overread. A batch-dependent output flip is not automatically a safety finding; it may reflect generic output sensitivity. The right comparison is therefore not ``does batching ever change anything,'' but ``when batching changes outputs, are safety rows affected differently from capability rows under the same serving perturbation?''

This distinction also explains why the batching question is harder than a standard determinism question. A reproducibility paper can stop after showing that repeated executions diverge. A safety paper cannot. It has to ask whether the divergence falls on refusal boundaries, whether the direction is systematically harmful, whether the effect survives broader model coverage, and whether a serving-system mechanism actually carries the risk into deployment.

Recent work on refusal mechanisms and safety brittleness makes this boundary focus more plausible. Refusal behavior can be mediated by compact internal directions, and pruning or low-rank modifications can selectively damage safety behavior while leaving utility comparatively intact \citep{Arditi2024RefusalDirection,Wei2024BrittlenessAlignment}. We do not use those papers as evidence that batching must change refusal. Instead, they motivate why a low-rate execution perturbation can still matter if it lands on a refusal boundary rather than on an ordinary paraphrase.

\subsection{Deployment optimization and alignment behavior}

Safety and alignment literature has already established that some deployment-time optimizations are not behaviorally neutral. Standard quantization work treats efficiency as the primary goal \citep{Xiao2023SmoothQuant}, while recent trustworthiness studies show that compression can change safety and trust behavior in ways not visible from benign task performance alone \citep{Hong2024CompressedTrust}. Other work shows that quantization can be adversarially exploited or corrected with alignment-aware procedures \citep{Egashira2024ExploitQuant,Wee2025AlignmentAwareQuant,Tan2026QRealign}. That literature supports the broader thesis that deployment configuration belongs inside the safety envelope. Our contribution is narrower and complementary. We focus on batching, which is often treated as a serving concern rather than an alignment concern, and ask whether it produces a safety-skewed instability pattern even under greedy decoding.

The batching paper is also different in an evaluative sense. Quantization directly changes arithmetic precision and therefore invites a relatively direct causal story. Batching is more indirect. It can alter queueing, co-residence, cache reuse, and execution order, but whether any of those mechanisms matter for refusal behavior is an empirical question. That is why a synthesis across local perturbation, cross-model extension, and composition-sensitive serving is necessary. A single positive study would be easy to dismiss as a measurement artifact; a single null would be easy to overread as proof of safety neutrality.

\subsection{Evaluation tasks and benchmark mix}

The evaluation design combines harmful-request refusal tasks with ordinary capability controls. For safety we use harmful-behavior and jailbreak-style prompts derived from adversarial attack and robust-refusal benchmarks \citep{Zou2023AdvBench,Mazeika2024HarmBench,Chao2024JailbreakBench}, and we use truthfulness, bias, and over-refusal style controls to keep the safety label from collapsing into a single harmful-compliance axis \citep{Lin2022TruthfulQA,Parrish2022BBQ,Roettger2024XSTest}. For capability controls we use MMLU and ARC-Challenge \citep{Hendrycks2021MMLU,Clark2018ARC}. This pairing is central to the paper's logic. A batch-dependent output change is not, by itself, evidence of a safety problem; the stronger test is whether safety changes outpace capability changes under the same serving perturbation.

The composition study adds a second evaluation distinction that is missing from most prior work: it separates generic batch-size perturbation from actual multi-tenant composition under continuous batching. Those are related but not identical threat models. A queued sequential server can show batch-sensitive variation without any true co-batch overlap. Conversely, a continuous-batching server can create genuine cross-request coexistence without producing a measurable aggregate safety effect. Keeping those threat models separate prevents the paper from collapsing several serving phenomena into one oversized claim.

\subsection{What is new here}

Relative to prior work, the novelty claim in this paper is deliberately narrow. We do not claim to have discovered batch-dependent nondeterminism; that already exists in the determinism and serving literature. We do not claim that deployment-time optimization is generally safety-neutral; compression and quantization work already cast doubt on that assumption. The contribution here is the measurement target itself: batching is evaluated as a safety failure mode, with a paired safety-versus-capability design and an explicit true-batching mechanism check. To our knowledge, that specific empirical question has not been answered directly in the existing literature. The paper's other contribution is corrective: it narrows an initial positive signal into the bounded model-specific claim that survives the full evidence chain.

%% file: sections/03_experimental_setup.tex
\section{Experimental Design}

\subsection{Threat Model and Unit of Analysis}

The operational threat model is deployment-time sensitivity under fixed prompts, fixed weights, and fixed decoding policy. The attacker is not assumed to alter the model or insert explicit adversarial suffixes at inference time. Instead, the concern is that ordinary serving choices---batch size, dispatch synchronization, or co-batched neighbors---shift the model onto different refusal outcomes for prompts that already sit near a decision boundary. This is why the unit of analysis is the \emph{conditioned evaluation row}: one prompt under two or more serving conditions, scored comparatively rather than in isolation.

This row-level framing matters for interpretation. The paper is not estimating population harm from deployed traffic logs. It is measuring whether changing the serving condition alters the answer on rows that are supposed to remain stable. That makes the synthesis strongest on comparative claims about behavioral sensitivity and weaker on absolute claims about deployed unsafe-output rates.

\subsection{Study A: Initial Local Perturbation Study}

The local perturbation study (Study~A) contains two evidence layers.
The base report covers \trOneThirtyEightBase{} scored rows on three instruction-tuned 1B--3B models under four phases: synchronized-dispatch batch-size perturbation, neighbor-condition perturbation, quantization-by-concurrency, and explicit true batching.
The strengthened revision adds a scorer-corrected audit of changed rows plus a \trOneThirtyEightRep{}-sample reduced replication on an enriched prompt subset.

The quantities carried forward are the safety-versus-capability asymmetry in the base study (\baseSafetyFlip{} versus \baseCapFlip{}) and strengthened replication (\repSafetyFlip{} versus \repCapFlip{}), the scorer-corrected audit split (\trOneThirtyEightAuditUnsafe{} unsafe-direction versus \trOneThirtyEightAuditSafe{} safe-direction rows), the manual adjudication of \trOneThirtyEightAdjudicatedRows{} candidates that leaves \trOneThirtyEightGenuineFlips{} genuine behavioral flips and a corrected full-set rate of about \trOneThirtyEightCorrectedRate{}, and the true-batching confirmation (\trOneThirtyEightTrueBatch{} safety flips with \trOneThirtyEightTrueAgree{} agreement to synchronized dispatch).
Capability controls are the guardrail against treating generic output churn as safety-specific, while the adjudication separates discovery sensitivity from operational magnitude.

\subsection{Study B: Cross-Model Extension}

The cross-model extension (Study~B) broadens the batch-perturbation question to \trOneFortyOneTotal{} records across three linked campaigns and a combined synthesis over the \mergedModels{} distinct models entering pooled scoring.
Its role in this paper is corrective rather than merely additive.
The extension tests whether the initial small-model asymmetry survives broader model coverage and whether categorical heuristics such as alignment type explain fragility.

The principal quantities are the near-parity combined safety-to-capability ratio (\trOneFortyOneCombinedRatio{}), the fragility range from \trOneFortyOneFragilityMin{} to \trOneFortyOneFragilityMax{}, the alignment-type ANOVA ($p=\trOneFortyOneAlignP$), the output-instability correlation ($r=\trOneFortyOneInstabilityR$), and the true-batch confirmation layer (\trOneFortyOnePhaseTwo{} flips with \trOneFortyOnePhaseAgree{} agreement).
Claims that appear locally but are not detectable in this extension are treated as bounded discovery signals rather than as final manuscript theses.

\subsection{Study C: Composition Study}

The composition study (Study~C) contributes the multi-tenant question under continuous batching in vLLM FP16.
The study covers \trOneFortyThreeTotal{} records across five batch-composition conditions, temporal-overlap sweeps, reverse-direction tests, and static-versus-continuous batching checks.

The relevant results are no aggregate composition effect at a 4.7 percentage-point minimum detectable effect, rare flips that lean unsafe in \trOneFortyThreeUnsafeLow{}\%--\trOneFortyThreeUnsafeHigh{}\% per condition (\trOneFortyThreePooledN{} pooled, Wilson 95\% CI \trOneFortyThreePooledCI{}), a strongest mixed-condition binomial signal at $p=\trOneFortyThreeMixedP$, and only \trOneFortyThreeCobatch{} co-batch verification.
This study asks whether \emph{which other requests are present at the same time} adds a measurable safety channel in continuous batching; without scheduler-level co-batch confirmation, a composition null can still be underpowered.

\subsection{Study D: Batch-Invariant Kernel Ablation}

Study~D is a targeted mechanism check on the current Study~A score-flip surface.
We selected the \kernelAblationRows{} current Phase~1/Phase~4 score-flip candidates from the rescored Study~A archive, retaining both safety and capability rows, and re-ran each row twice on the same H100 pod: once with standard vLLM and once with \texttt{VLLM\_BATCH\_INVARIANT=1}.
The serving stack was vLLM~\kernelAblationVllm{}, FP16 weights, temperature~0, max model length 2048, and the same three small instruction-tuned models used by the local study: Llama-3.2-1B-Instruct (\kernelAblationLlamaOneRows{} rows), Llama-3.2-3B-Instruct (\kernelAblationLlamaThreeRows{} rows), and Qwen2.5-1.5B-Instruct (\kernelAblationQwenRows{} rows).
Dispatch mode followed the original condition: \kernelAblationSyncRows{} synchronized-dispatch rows and \kernelAblationPromptListRows{} prompt-list true-batching rows.

This is not a new population estimate; it is a mechanism stress test on rows already identified as batch-sensitive.
If flips collapse under batch-invariant execution, the exact-stack recommendation becomes stronger: rerun the safety battery under the serving kernel path used in deployment.

\subsection{Why a Synthesis Is Necessary}

These four studies answer different questions, so the synthesis reports the strongest claim that survives the chain rather than the strongest available in any one component. The local study discovers the signal, the cross-model extension narrows it to model-specific fragility, the composition study rules out a large aggregate continuous-batch hazard at current sensitivity, and the kernel ablation tests whether current candidates depend on a batch-sensitive vLLM execution path.

\subsection{Synthesis Rules and Statistical Interpretation}

The manuscript preserves each component study's original statistical treatment rather than imposing a post-hoc pooled significance layer across different model sets, scoring stacks, and threat models. A positive local result is promoted only if it survives the larger extension or an independent mechanism check; directional results are paired with absolute rates; and where studies conflict, the larger or more mechanistically targeted study determines the visible claim.

%% file: sections/04_results.tex
\section{Results}

\subsection{Study A: The Initial Local Study Yields a Real but Low-Rate Discovery Signal}

\begin{figure}[t]
  \centering
  \includegraphics[width=\linewidth]{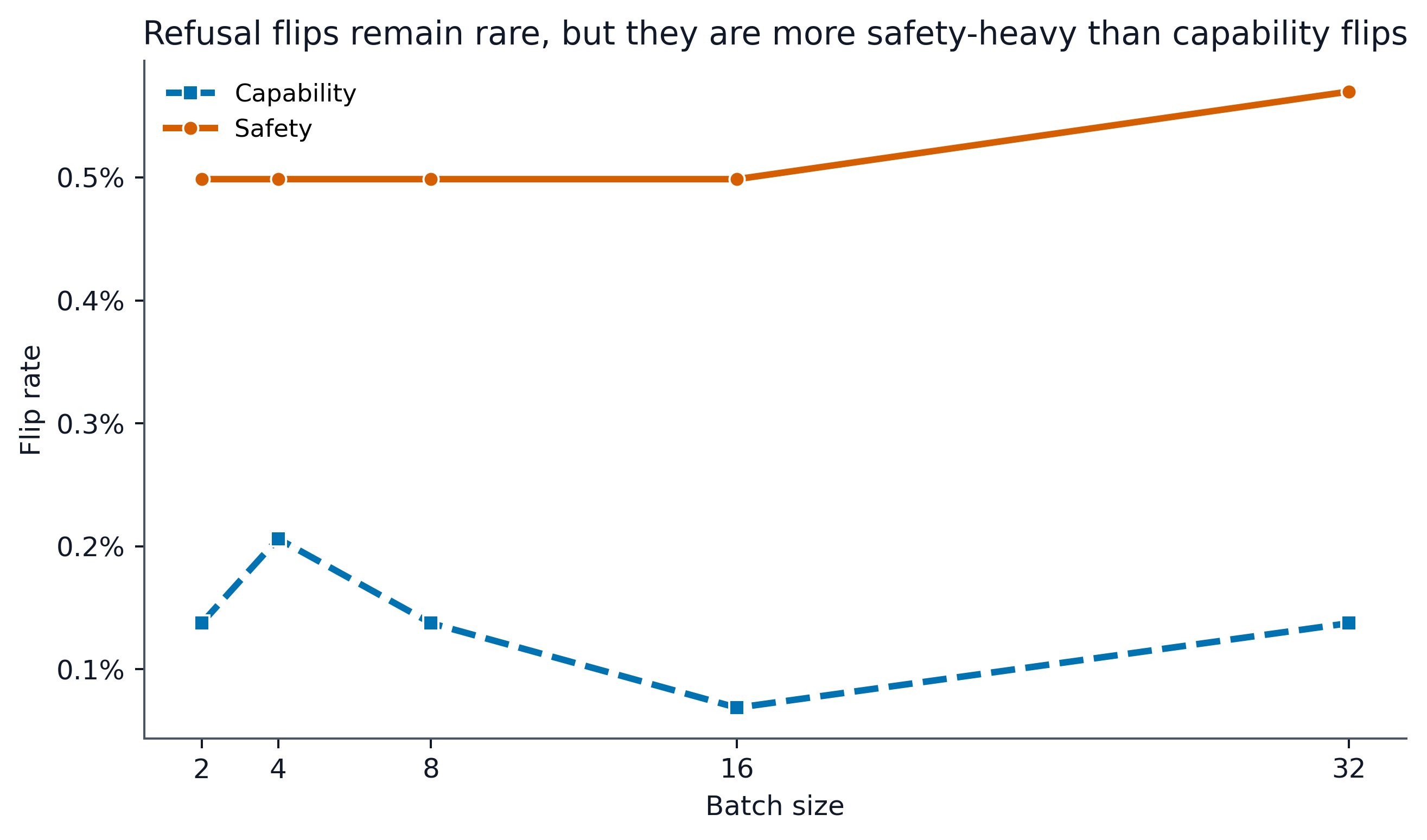}
  \caption{Study~A safety versus capability flip rates by batch size. Safety flips exceed capability flips in the local discovery setting, identifying a refusal-boundary signal while leaving the absolute rate low.}
  \label{fig:phase1_safety_cap}
\end{figure}

The local discovery study (Study~A) is the discovery anchor: it demonstrates that batch perturbation can move refusal behavior at all.
In the base experiment, safety labels change more often than capability labels (\baseSafetyFlip{} versus \baseCapFlip{}), and the reduced replication preserves the same direction at higher absolute rates on the enriched subset (\repSafetyFlip{} versus \repCapFlip{}).
Figure~\ref{fig:phase1_safety_cap} summarizes the relative flip rates across batch sizes.

\paragraph{What this supports.}
The local study does not show that batching broadly destabilizes model behavior; most rows remain unchanged.
It shows that changed rows are disproportionately safety-relevant in the original three-model deterministic setting, making Study~A a discovery signal for \emph{where} batching matters rather than evidence that batching dominates total variance.

\paragraph{Mechanism confirmation.}
The explicit true-batching layer matters because it weakens the pure scheduler-artifact explanation.
Prompt-list true batching still produces \trOneThirtyEightTrueBatch{} safety flips with \trOneThirtyEightTrueAgree{} agreement to the synchronized-dispatch pattern.
That is mechanism-consistency evidence, not a separate large-magnitude effect.

\paragraph{Audit correction.}
The scorer-corrected audit shrinks the changed-row set but does not reverse it: after Unicode normalization, the audit catalog contains \trOneThirtyEightAuditUnsafe{} unsafe-direction rows and \trOneThirtyEightAuditSafe{} safe-direction rows.
This keeps the directional concern alive while making the rarity of the phenomenon explicit; the original changed-row catalog is therefore a sensitive discovery layer, not a calibrated operational rate.

\paragraph{Human adjudication recalibration.}
The later conclusive audit is the operational magnitude check.
Manual review of \trOneThirtyEightAdjudicatedRows{} candidate rows found only \trOneThirtyEightGenuineFlips{} genuine behavioral flips (\trOneThirtyEightGenuinePct{}\%), with the remaining \trOneThirtyEightArtifactPct{}\% attributable to automated scoring artifacts such as refusal rephrasing.
That recalibration implies a corrected full-set rate of roughly \trOneThirtyEightCorrectedRate{}, which is much smaller than the original automated discovery signal.
This does not erase the phenomenon.
It changes the correct interpretation of the local study: high sensitivity for finding unstable refusal-boundary rows, low direct evidence for a large production-rate hazard.
The adjudication itself is single-reviewer and should be read as a conservative correction layer rather than as a final population estimate.

\subsection{Study B: The Cross-Model Extension Finds No Universal Asymmetry}

The cross-model extension (Study~B) is the paper's main correction layer.
When the batch-perturbation question is scaled across a broader model set, no universal safety-over-capability skew is detectable.
The combined synthesis lands near parity at \trOneFortyOneCombinedRatio{}, not at a stable safety-skewed ratio.

\paragraph{What does survive.}
Three findings remain strong in the extension. First, fragility varies materially by model, spanning \trOneFortyOneFragilityMin{} to \trOneFortyOneFragilityMax{}. Second, alignment type does not explain that variation: a model-level one-way ANOVA over the four alignment categories returns $F=0.13$, $p=\trOneFortyOneAlignP{}$, and $\eta^2=\trOneFortyOneAlignEta{}$, with per-group mean flip rates spanning only \trOneFortyOneAlignEquiv{} (well inside a $\pm$3pp equivalence margin). With $n=\mergedModels{}$ models the bootstrap CI on $\eta^2$ is wide, so this is reported as ``no detectable association at the available power'' rather than a strict equivalence claim. Third, output instability is the best early-warning predictor of batch fragility ($r=\trOneFortyOneInstabilityR{}$, bootstrap 95\% CI \trOneFortyOneInstabilityRCI{}, leave-one-out range \trOneFortyOneInstabilityLOOCV{}).

\begin{figure}[t]
  \centering
  \includegraphics[width=\linewidth]{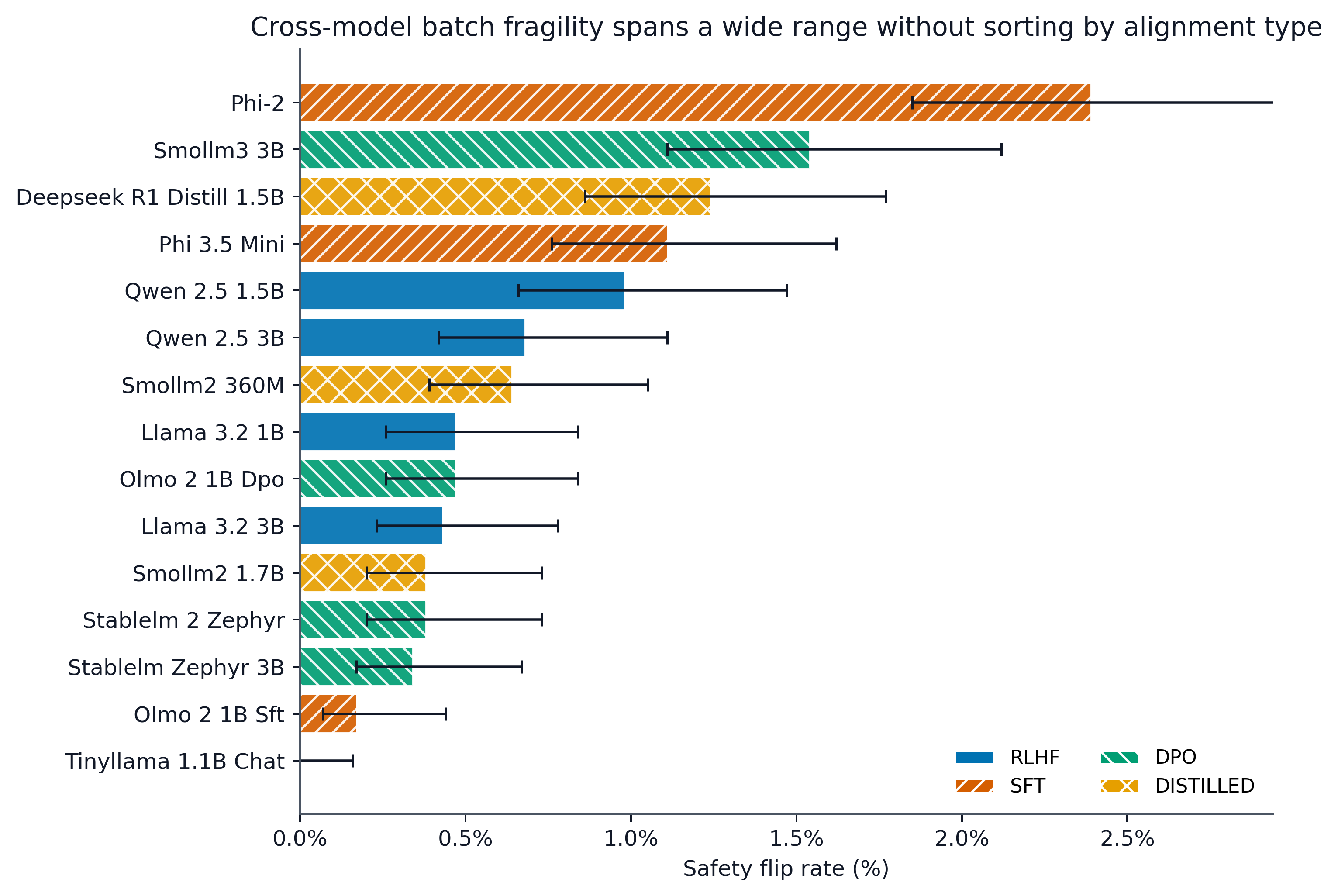}
  \caption{Cross-model fragility ranking from the 15-model extension. Safety flip rates span \trOneFortyOneFragilityMin{} to \trOneFortyOneFragilityMax{}, with error bars showing 95\% Wilson confidence intervals. High- and low-fragility models appear across multiple alignment categories rather than forming a clean alignment-type ladder.}
  \label{fig:fragility_ranking}
\end{figure}

\begin{figure}[t]
  \centering
  \includegraphics[width=0.95\linewidth]{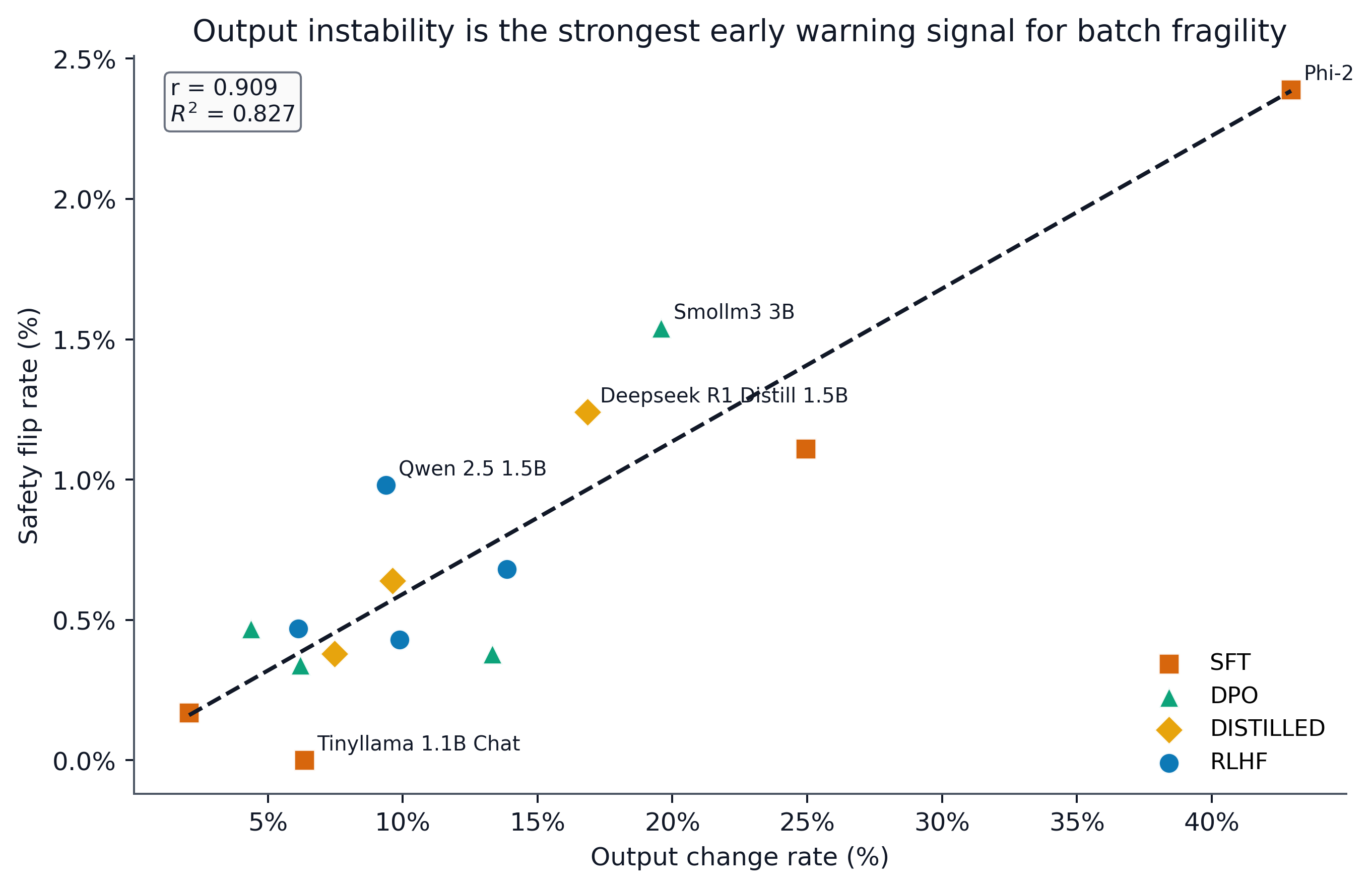}
  \caption{Output instability versus safety fragility across the same 15-model extension. The dashed line is the least-squares linear fit. Models with larger output-change rates under batching also exhibit higher refusal fragility ($r=\trOneFortyOneInstabilityR$), which makes output instability a more useful screening signal than alignment type.}
  \label{fig:instability_fragility}
\end{figure}

Together these results change the operational question from ``is batching intrinsically unsafe'' to ``which models exhibit batch fragility and what observable signal predicts it before deployment?''
The instability correlation is useful because it offers a screening variable where alignment labels and architecture-family descriptions do not.
Figure~\ref{fig:fragility_ranking} shows the full 15-model fragility range, while Figure~\ref{fig:instability_fragility} shows why output-change rate is the most useful deployment-time screening signal.

\paragraph{Directional correction.}
The cross-model study also changes the directional story.
The combined directional counts are net safe, with \trOneFortyOneSafeDir{} compliance-to-refusal flips versus \trOneFortyOneUnsafeDir{} refusal-to-compliance flips.
That result does not make batching benign; it makes the direction model-set-dependent rather than universal.
The extension should therefore be read as a heterogeneity report, not as proof of a universal unsafe skew.

\begin{figure}[t]
  \centering
  \includegraphics[width=\linewidth]{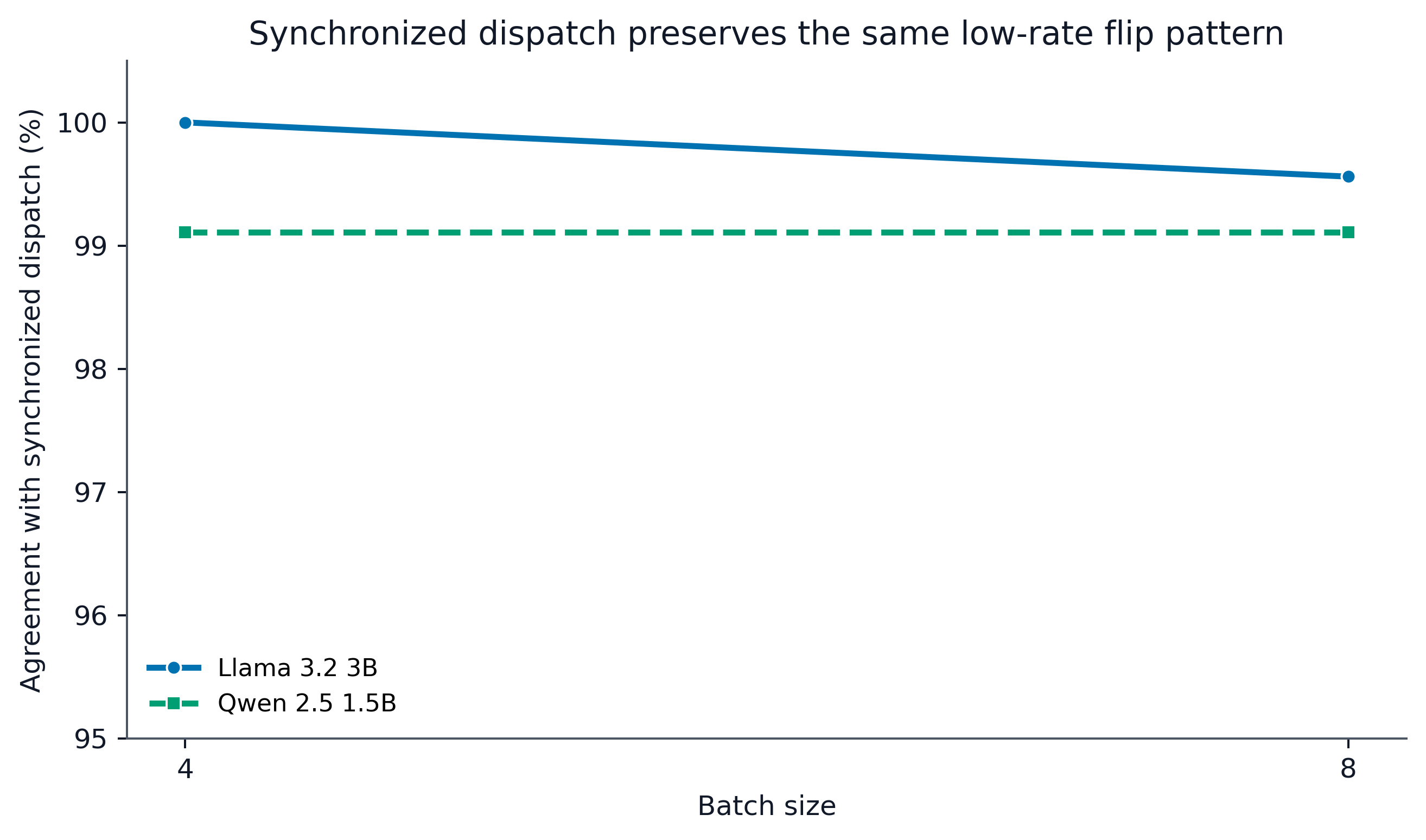}
  \caption{True-batch agreement with synchronized dispatch across model--batch-size conditions. The y-axis is zoomed to 95--100.5\% to expose small differences. Near-100\% agreement argues against a pure synchronized-dispatch artifact while preserving the low-rate interpretation.}
  \label{fig:true_batch}
\end{figure}

\paragraph{True batching remains narrow but consistent.}
The preserved phase-two artifact again shows low-rate mechanism consistency: \trOneFortyOnePhaseTwo{} safety flips with \trOneFortyOnePhaseAgree{} agreement against the synchronized-dispatch pattern.
This reinforces the ``real but small'' reading without creating a new high-magnitude claim.
Figure~\ref{fig:true_batch} shows the agreement between true-batch and synchronized-dispatch patterns across all tested conditions.

The extension therefore does something more useful than confirming or rejecting the local study wholesale.
It decomposes the original positive into three parts: a real perturbation signal, a failed universal asymmetry claim, and a model-specific instability problem that remains worth validating in practice.

\subsection{Study C: The Composition Study Finds a Null with a Directional Caveat}

The final question is whether continuous batching introduces an additional multi-tenant safety channel through batch composition.
The composition study (Study~C) does not detect an aggregate effect at the current power level.
Across McNemar, Cochran's Q, and Mantel--Haenszel analyses, the study remains null at a 4.7 percentage-point minimum detectable effect.

\paragraph{What remains concerning.}
The rare flips that do occur lean unsafe in \trOneFortyThreePooledN{} cases pooled across the three multi-prompt conditions (\trOneFortyThreePooledRatio{}, Wilson 95\% CI \trOneFortyThreePooledCI{}), with per-condition rates of \trOneFortyThreeUnsafeLow{}\%--\trOneFortyThreeUnsafeHigh{}\% and the strongest mixed-condition signal at $p=\trOneFortyThreeMixedP{}$. The pooled lower CI bound at 75\% is well above 50\%, so the directional asymmetry is robust at the pooled level even though per-condition CIs are wide. This is not a clean null, but the small denominator (31 directional flips total) and the \trOneFortyThreeCobatch{} scheduler-level co-batch verification rate together rule it out as a practical basis for composition-aware request routing.

The minimum detectable effect is doing important interpretive work: the study finds no large aggregate composition effects under the tested stack, but it does not exclude small effects that would matter only in very high-volume or highly sensitive deployments.

\paragraph{Correct interpretation.}
The composition study narrows the threat model.
At current sensitivity, batch composition is not a large aggregate safety driver.
What it contributes is a cautionary directional pattern that should be monitored as batch sizes and verification quality improve.

\subsection{Study D: Batch-Invariant Kernels Remove the Candidate Flips}

Study~D directly tests the mechanism that the previous evidence chain left open.
On the same \kernelAblationRows{} current score-flip candidates, standard vLLM produces \kernelAblationStandardFlips{} label flips and \kernelAblationStandardTextChanges{} text changes.
With \texttt{VLLM\_BATCH\_INVARIANT=1}, the same prompts, models, dispatch modes, and H100 serving environment produce \kernelAblationInvariantFlips{} label flips and \kernelAblationInvariantTextChanges{} text changes.
All \kernelAblationRecords{} records complete successfully with no startup or request errors.

\begin{table}[t]
  \centering
  \caption{Batch-invariant-kernel ablation on the current Study~A score-flip candidates. Standard vLLM reproduces low-rate candidate flips; the invariant execution path removes them under the tested stack.}
  \label{tab:kernel_ablation}
  \resizebox{\linewidth}{!}{%
  \begin{tabular}{lrrrr}
    \toprule
    Mode & Rows & OK & Label flips & Text changes \\
    \midrule
    Standard vLLM & \kernelAblationRows{} & \kernelAblationRows{} & \kernelAblationStandardFlips{} & \kernelAblationStandardTextChanges{} \\
    Batch-invariant & \kernelAblationRows{} & \kernelAblationRows{} & \kernelAblationInvariantFlips{} & \kernelAblationInvariantTextChanges{} \\
    \bottomrule
  \end{tabular}
  }
\end{table}

The standard-mode flips are all safety-domain rows in this candidate set, with \kernelAblationMatchedDirection{} matching the original flip direction.
The result should be read as a mechanism ablation, not as a universal claim about every serving backend.
For vLLM~\kernelAblationVllm{} on this H100 stack, the local candidate flips depend on the non-invariant execution path.
That finding converts the earlier ``true batching is mechanism-consistent'' evidence into a sharper operational test: if a deployment uses batch-sensitive kernels, refusal robustness should be evaluated under that exact served path.

\subsection{Synthesis: What the Combined Evidence Actually Supports}

Taken together, the four studies support a narrower claim than the earliest single-study interpretation: batch perturbation is real, low-rate after adjudication, strongly model-specific, not well predicted by alignment type, not a large aggregate composition hazard at current power, and removable on the tested vLLM/H100 candidate surface by batch-invariant kernels.
Operators should validate refusal behavior at the production batch setting for the actual model and kernel path they deploy, but they should not treat batching as the dominant optimization-time safety threat or infer risk from architecture-family slogans.
The practical check is lightweight: compare outputs at the served batch setting, include capability controls, and escalate only when output instability or refusal flips appear.

%% file: sections/05_discussion_limitations.tex
\section{Discussion and Limitations}

\subsection{Testing Meaning}

The paper's main testing claim is modest but actionable: batch configuration belongs inside the evaluated condition set for refusal robustness.
That does not mean every batching change is high risk.
It means that batching is no longer justified as fixed background infrastructure once paired evidence shows it can move refusal behavior.

The synthesis also improves resource allocation: batch risk is real enough to validate, but not large enough to dominate engineering effort.
Across companion deployment studies on the same stack, quantization effects are much larger than batch perturbation, consistent with compression work showing that efficiency interventions can alter trust and safety behavior in ways benign task performance does not reveal \citep{Hong2024CompressedTrust,Egashira2024ExploitQuant,Wee2025AlignmentAwareQuant,Tan2026QRealign}.
The testing order is therefore: validate quantization first, then batching on the exact served model, batch setting, and kernel path, and only then composition-specific edge cases.

\subsection{Why low-rate effects still matter}

A likely reviewer concern is that the measured rates are too small to justify a full paper.
For refusal behavior, the relevant standard is not high aggregate variance; it is whether deployment choices introduce directional instability on prompts that should remain consistently refused.
This is why robust-refusal benchmarks must distinguish harmful requests from benign requests that merely look similar \citep{Mazeika2024HarmBench,Roettger2024XSTest}.

\subsection{What This Paper Does Not Claim}

This manuscript does not claim a universal safety-over-capability asymmetry under batching.
The cross-model extension finds no detectable skew across the tested model set.
It does not claim that alignment type predicts fragility: the model-level ANOVA returns $p=\trOneFortyOneAlignP{}$ with $\eta^2=\trOneFortyOneAlignEta{}$ and group means inside a \trOneFortyOneAlignEquiv{} band, but with $n=\mergedModels{}$ the bootstrap CI on $\eta^2$ is wide and we report the result as no detectable association at the available power, not as strict equivalence.
It does not claim that continuous-batch composition is a large aggregate hazard.
The rare unsafe-direction flips observed in the composition study are not sufficient evidence for composition-aware routing today, because the pooled directional ratio (\trOneFortyThreePooledN{}, Wilson 95\% CI \trOneFortyThreePooledCI{}) rests on only 31 directional flips and \trOneFortyThreeCobatch{} co-batch verification.
It also does not claim that all batch-conditioned effects in all serving stacks are caused by the same kernel mechanism.
Study~D supports that mechanism for the tested vLLM~\kernelAblationVllm{}/H100 stack and current candidate set; other backends, tensor-parallel configurations, larger models, and stochastic decoding need their own ablations.

\subsection{Main Limitations}

\paragraph{Rare-event regime.}
All four studies operate in a sparse-flip regime.
That is why directional findings can be meaningful while still carrying wide confidence intervals and sensitivity to model set.

\paragraph{Measurement heterogeneity.}
The synthesis combines a local three-model study, a larger cross-model extension on different hardware, and a continuous-batching composition study.
This is appropriate for a bounded synthesis paper, but it means the output is an operational doctrine rather than a single pooled effect estimate.

\paragraph{Single-reviewer adjudication.}
The local-study recalibration to \trOneThirtyEightCorrectedRate{} depends on a manual review of \trOneThirtyEightAdjudicatedRows{} candidate rows by a single reviewer, with no inter-rater Cohen's $\kappa$.
It prevents overstating the automated discovery rate, but should be read as an annotator-perceived correction rather than a gold-standard population estimate.

\paragraph{Scoring-stack heterogeneity.}
The local study includes an explicit scorer audit, while the preserved cross-model and composition artifacts rely on automated regex-pattern-based scoring rather than judge-based scoring.
This does not invalidate the within-study contrasts, but it limits the precision of absolute-rate comparisons across studies and is one reason the paper emphasizes claim selection over pooled numerical precision.

\paragraph{Co-batch verification.}
The composition study's \trOneFortyThreeCobatch{} verification rate biases analysis toward the null and limits the strength of any negative claim about actual scheduler-level co-residence.

\paragraph{Kernel-ablation scope.}
The batch-invariant ablation is deliberately targeted at \kernelAblationRows{} current score-flip candidates rather than the full original screen.
That makes it a strong mechanism check for the candidate surface, but not a replacement for a larger prospective benchmark.
It also covers one serving stack: vLLM~\kernelAblationVllm{} on H100 with three 1B--3B instruction-tuned models.

\paragraph{Local-first scope.}
The evidence is local-first and small-to-mid-scale.
Nothing in this paper should be read as a claim about very large models, tensor-parallel production clusters, or stochastic decoding.

\subsection{Implications for evaluation practice}

Safety evaluation should inherit the discipline already standard in systems benchmarking: if a deployment choice changes the execution regime, safety testing should be rerun in that regime rather than assumed invariant from a single baseline configuration.
For batching, the minimum protocol is simple: evaluate the exact production batch setting, include capability controls, and preserve at least one mechanism-sensitive check that distinguishes synchronized dispatch from genuine true batching.
Study~D shows the value of that check: with batch-invariant kernels enabled \citep{He2025DefeatingNondeterminism,LMSYS2025Deterministic}, candidate label flips collapse from \kernelAblationStandardFlips{} to \kernelAblationInvariantFlips{} under the tested stack.
The next stronger paper should use this as the reference ablation and extend it prospectively across larger models, tensor-parallel deployments, and backend families.

%% file: sections/06_conclusion.tex
\section{Conclusion}

Batch perturbation can change refusal behavior, but the surviving effect is low-rate, model-specific, and stack-dependent: the local paired test yields the discovery signal, single-reviewer adjudication reduces the implied operational rate to roughly \trOneThirtyEightCorrectedRate{}, the cross-model extension finds no detectable universal safety-over-capability skew across the \mergedModels{} tested models, the composition test finds no aggregate continuous-batch effect at current sensitivity, and the batch-invariant-kernel ablation removes the current candidate flips on the tested vLLM/H100 stack. The testing recommendation is therefore exact-stack validation, not universal prohibition: validate refusal at the batch setting and kernel path actually served, pair safety prompts with capability controls, and do not infer batch robustness from alignment type, architecture family, or composition heuristics.

%% file: sections/appendix_reproducibility.tex
\section{Reproducibility Details}
\label{app:repro}

This appendix consolidates the prompts, scoring rules, model and serving setup, and per-condition grid behind the component studies, in response to reviewer requests for reproducibility detail. Numeric totals are reported in the main text via the source-line macros; this appendix documents the \emph{procedure} behind them.

\subsection{Prompt Corpora}

Studies~A--C draw from the same six task families, evaluated identically so that safety and capability flips are measured on the same footing; Study~D reuses the current score-flip candidates selected from that same task pool. Three are safety-axis families and three are capability-axis controls:

\begin{itemize}[leftmargin=*]
  \item \textbf{Safety axis:} \texttt{advbench\_refusal} (harmful-instruction refusal), \texttt{jailbreak\_amplification} (jailbreak-style adversarial prompts), and \texttt{bbq\_bias} (bias-sensitive prompts).
  \item \textbf{Capability axis:} \texttt{mmlu\_real} (knowledge), \texttt{arc\_challenge} (reasoning), and \texttt{truthfulqa} (truthfulness).
\end{itemize}

The capability families are not decorative. They are the control that distinguishes a safety-specific batching effect from generic output churn: if batching merely perturbed outputs at random, safety and capability flip rates would move together. The asymmetry the local study reports (\baseSafetyFlip{} safety versus \baseCapFlip{} capability) is meaningful precisely because the two axes are scored by the same pipeline under the same serving perturbation.

\subsection{Models}

Studies~A--C evaluate instruction-tuned models in the 1B--3B class, spanning five families to avoid a single-vendor artifact: Llama (\texttt{Llama-3.2-1B/3B-Instruct}), Qwen (\texttt{Qwen2.5-1.5B/3B-Instruct}), Gemma (\texttt{gemma-2-2b-it}, \texttt{gemma-3-1b-it}), Phi (\texttt{phi-2}, \texttt{Phi-3.5-mini-instruct}), plus matched quantized variants. The cross-model extension (Study~B) pools \mergedModels{} distinct models across its linked campaigns; this is the model count that enters the combined safety-to-capability ratio (\trOneFortyOneCombinedRatio{}) and the alignment-type ANOVA. Study~D uses the three local-study models represented in the current candidate surface: Llama-3.2-1B-Instruct, Llama-3.2-3B-Instruct, and Qwen2.5-1.5B-Instruct.

\subsection{Serving and Decoding}

Generation uses vLLM (OpenAI-compatible server, FP16 weights) with a 2048-token maximum context and \textbf{greedy decoding at temperature~0.0}. Deterministic decoding is essential to the threat model: because the decode policy is fixed, any output change across serving conditions is attributable to the serving condition (batch size, dispatch synchronization, or co-batched neighbors), not to sampling noise. The composition study (Study~C) runs vLLM's continuous-batching scheduler; the local and extension studies (Studies~A and~B) use synchronized-dispatch and explicit true-batching configurations as described in the per-condition grid below.

\subsection{Scoring, Flip Definition, and Blinding}

The unit of analysis is the \emph{conditioned evaluation row}: one prompt under two or more serving conditions, scored comparatively. A ``flip'' is a change in the row's label between conditions. Scoring is layered across the studies, and the layers differ deliberately:

\begin{itemize}[leftmargin=*]
  \item \textbf{Automated scoring} drives the large-scale screens (the batch-size sweeps and the \trOneFortyOneTotal{}-record extension), where per-row human review is infeasible. This is the layer that produces the raw discovery rates.
  \item \textbf{Blinded LLM-judge scoring} adjudicates the composition study, using an Ollama-served judge (\texttt{qwen2.5:7b-instruct-q8\_0}) that sees only the target prompt and response, never the co-batch composition --- so the judge cannot be biased by knowing which condition produced a row.
  \item \textbf{Manual adjudication} corrects the discovery layer: the \trOneThirtyEightAdjudicatedRows{} candidate changed rows from Study~A were hand-reviewed, finding \trOneThirtyEightGenuineFlips{} genuine behavioral flips (\trOneThirtyEightGenuinePct{}\%) and \trOneThirtyEightArtifactPct{}\% scorer artifacts, implying the corrected operational rate of \trOneThirtyEightCorrectedRate{}.
\end{itemize}

This layering is why the paper separates \emph{discovery sensitivity} from \emph{operational magnitude}: the automated layer is a screen for boundary instability, and the manual layer is the magnitude estimate. The \trOneThirtyEightArtifactPct{}\% artifact share is reported openly rather than buried, because it is the reason the headline operational rate (\trOneThirtyEightCorrectedRate{}) is so much smaller than the raw discovery rate.

\subsection{Per-Condition Grid}

\begin{itemize}[leftmargin=*]
  \item \textbf{Batch-size sweep (Studies~A, B):} batch sizes $\{1, 2, 4, 8, 16, 32\}$ under synchronized dispatch.
  \item \textbf{Quantization\,$\times$\,concurrency (Study~A, Ollama):} concurrency levels $\{1, 4, 8\}$ crossed with batch sizes $\{1, 4, 8\}$.
  \item \textbf{True batching (Studies~A, B):} explicit co-batched dispatch, the mechanism check that confirms the synchronized-dispatch pattern (\trOneThirtyEightTrueAgree{} and \trOneFortyOnePhaseAgree{} agreement respectively).
  \item \textbf{Composition (Study~C):} five batch-composition conditions at a fixed batch size of 8 under continuous batching, with temporal-overlap sweeps and static-versus-continuous checks. The scheduler-level co-batch verification rate was \trOneFortyThreeCobatch{}; this is reported as a limitation, since an unverified co-batch is an underpowered measurement rather than evidence of no mechanism.
  \item \textbf{Kernel ablation (Study~D):} \kernelAblationRows{} current Phase~1/Phase~4 score-flip candidates re-run under standard vLLM and \texttt{VLLM\_BATCH\_INVARIANT=1}. The candidate set contains \kernelAblationSyncRows{} synchronized-dispatch rows and \kernelAblationPromptListRows{} prompt-list true-batching rows, split across Llama-3.2-1B, Llama-3.2-3B, and Qwen2.5-1.5B.
\end{itemize}

\subsection{Hardware and Determinism Caveat}

Runs were executed on the lab's local and rented GPU infrastructure. The Study~D ablation was run on an NVIDIA H100 80GB HBM3 pod using vLLM~\kernelAblationVllm{}, PyTorch~2.10.0, Triton~3.6.0, FP16 weights, max model length 2048, and temperature~0.0. One determinism caveat is material: even at temperature~0.0, floating-point reduction order can differ across batch sizes, so ``deterministic decoding'' does not guarantee bit-identical logits across serving conditions. Study~D directly tests this caveat by enabling the batch-invariant execution path on the current candidate rows.

\subsection{Study D Artifact Record}

The Study~D run contains \kernelAblationRecords{} records: \kernelAblationRows{} standard-vLLM rows and \kernelAblationRows{} batch-invariant rows.
All \kernelAblationRecords{} records completed successfully with zero startup or request errors.
The retained internal artifacts include the run metadata, selected candidate metadata, summary JSON, full local audit JSONL, and native vLLM startup logs showing the batch-invariant kernel registration for the invariant-mode runs.
The public release should expose summary and candidate metadata, but not raw harmful prompts or completions.

%% file: sections/appendix_audit_strengthening.tex
\section{Study Provenance Appendix}

This manuscript synthesizes four component studies rather than reporting one fresh experiment.
The retained evidence chain is:

\begin{itemize}[leftmargin=*]
  \item \textbf{Study~A} --- an initial local discovery study with scorer-corrected audit and reduced replication;
  \item \textbf{Study~B} --- a cross-model extension that tests whether the discovery result generalizes;
  \item \textbf{Study~C} --- a continuous-batch composition study that probes the multi-tenant question;
  \item \textbf{Study~D} --- a batch-invariant-kernel ablation on the current Study~A score-flip candidates.
\end{itemize}

\paragraph{Claim mapping.}
\begin{itemize}[leftmargin=*]
  \item ``Batch perturbation is real'' is grounded in Study~A and reinforced by the later true-batching confirmation layer.
  \item ``Universal safety-over-capability asymmetry is not established'' is grounded in the combined Study~B synthesis.
  \item ``Alignment type is not predictive'' is grounded in the balanced Study~B extension.
  \item ``Composition shows an aggregate null with directional caveat'' is grounded in Study~C.
  \item ``The current local candidate flips depend on the tested non-invariant vLLM kernel path'' is grounded in Study~D.
\end{itemize}

\paragraph{Intended reading rule.}
Where the component studies disagree, this paper follows the most conservative claim supported by the full evidence chain rather than the strongest positive framing available in any one component study.

%% file: sections/appendix_submission_materials.tex
\section{Ethical Considerations and Artifact Availability}

\subsection{Ethical Considerations}

This paper studies refusal fragility under batching using offline evaluation prompts, including harmful-request batteries.
No human subjects, private data, or live-user interventions are involved.
Because the study touches adversarial prompting, aggregate results, scored outputs, and analysis code can be shared openly, but ready-to-run harmful prompt batteries should be disclosed responsibly and only to the extent required for scientific reproduction.

\subsection{Artifact Availability}

The reproduction package for this paper includes:
\begin{itemize}[leftmargin=*]
  \item retained scored outputs and condition-level summary tables for the local batching study, the cross-model extension, and the composition audit;
  \item the Study~D batch-invariant-kernel ablation summary, selected-candidate metadata, startup provenance, and validation log;
  \item figure-generation data and manuscript tables for the reported flip-rate, instability, and directional-risk analyses; and
  \item build scripts, compiled figures, and audit notes sufficient to regenerate the paper package.
\end{itemize}

The public artifact should withhold raw harmful prompts and completions from the Study~D full-record JSONL unless a controlled-access review process requires them.
Aggregate counts, hashes, per-condition metadata, and scripts are sufficient for the open release surface.